\documentclass[11pt,a4paper]{article}
\usepackage[hyperref]{acl2021}
\usepackage{times}
\usepackage{latexsym}

\usepackage{microtype}

\usepackage{multirow}
\usepackage{booktabs}
\usepackage{courier}
\usepackage{amsmath}
\usepackage{subfig}
\usepackage{placeins}
\usepackage{tikz}
\usepackage{hyperref}

\aclfinalcopy 

\newcommand{\bee}{\raisebox{-2pt}{\includegraphics[width=0.15in]{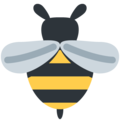}}}
\newcommand{\idiap}{\raisebox{-2pt}{\includegraphics[width=0.15in]{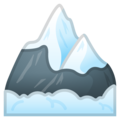}}}

\makeatletter
\newcommand{\printfnsymbol}[1]{%
  \textsuperscript{\@fnsymbol{#1}}%
}
\makeatother

\title{Does My Representation Capture $X$? Probe-Ably}

\author{Deborah Ferreira\thanks{\ \ Equal contribution, presented in alphabetical order.}$~^{\bee}$, Julia Rozanova\printfnsymbol{1}$^{\bee}$, Mokanarangan Thayaparan\printfnsymbol{1}$^{\bee}$,\\ \textbf{Marco Valentino\printfnsymbol{1}$^{\bee}$, Andr\'e Freitas$^{\bee \idiap}$} \\  Department of Computer Science, University of Manchester, United Kingdom$^{\bee}$ \\  Idiap Research Institute, Switzerland $^{\idiap}$ \\ {\tt \{firstname.surname\}} {\tt @manchester.ac.uk} \\}

\begin{document}
\maketitle
\begin{abstract}
Probing (or \emph{diagnostic classification}) has become a popular strategy for investigating
whether a given set of intermediate features is present in the representations of neural models. 
Probing studies may have misleading results, but various recent works have suggested
more reliable methodologies that compensate for the possible pitfalls of probing. 
However, these best practices are numerous and fast-evolving. 
To simplify the process of running a set of probing experiments in line with suggested methodologies, 
we introduce \textbf{Probe-Ably:} an extendable probing framework
which supports and automates the application of probing methods
to the user's inputs. 
\end{abstract}

\section{Introduction}
Recent interest in investigating the intermediate features present
in neural models' representations has led to the use of structural analysis methods
such as \emph{probing}.

At its simplest, \textbf{probing}\footnote{The term ``probing" has also been used describe stress-test 
style analyses,
but we mean ``probing" in the sense of \emph{diagnostic classification} as in
\cite{alain-bengio, pimentel2020information}.} 
is the training of an external classifier model (a ``probe")
to determine the extent to which a set of auxiliary target feature labels can be 
predicted from the internal model representations. For example, probing studies have been carried out to determine whether word and sentence representations
generated by models such as BERT \cite{devlin2019bert} capture intermediate syntactic and
semantic features such as parts of speech and dependency labels \cite{hewitt2019structural, tenney2018what} and lexical relations \cite{vulic2020probing}. 

Various problems can arise when performing probing experiments \cite{hewitt-liang}, such as achieving a high probing accuracy without being due to a high mutual information
between the representation and the auxiliary task labels.
This has prompted much recent work on establishing more reliable methodologies
for probing \cite{hewitt-liang,voita2020information, pimentel2020information, pimentel2020pareto}. 

These approaches introduce various steps such as controlling and varying 
model complexity and structure, including randomized control tasks 
and incorporating  more informative metrics such as selectivity
\cite{hewitt-liang} and minimum description length 
\cite{voita2020information}.

To make these methods more accessible and quick to implement for any user wishing to probe the 
representations of their neural models in line with the evolving suggested methodologies,
we introduce \textbf{Probe-Ably:} an extendable probing framework
which supports and automates the application of suggested best practices for probing studies.

\section{Probe-Ably}
\begin{figure*}
        \centering
        \includegraphics[width=\textwidth]{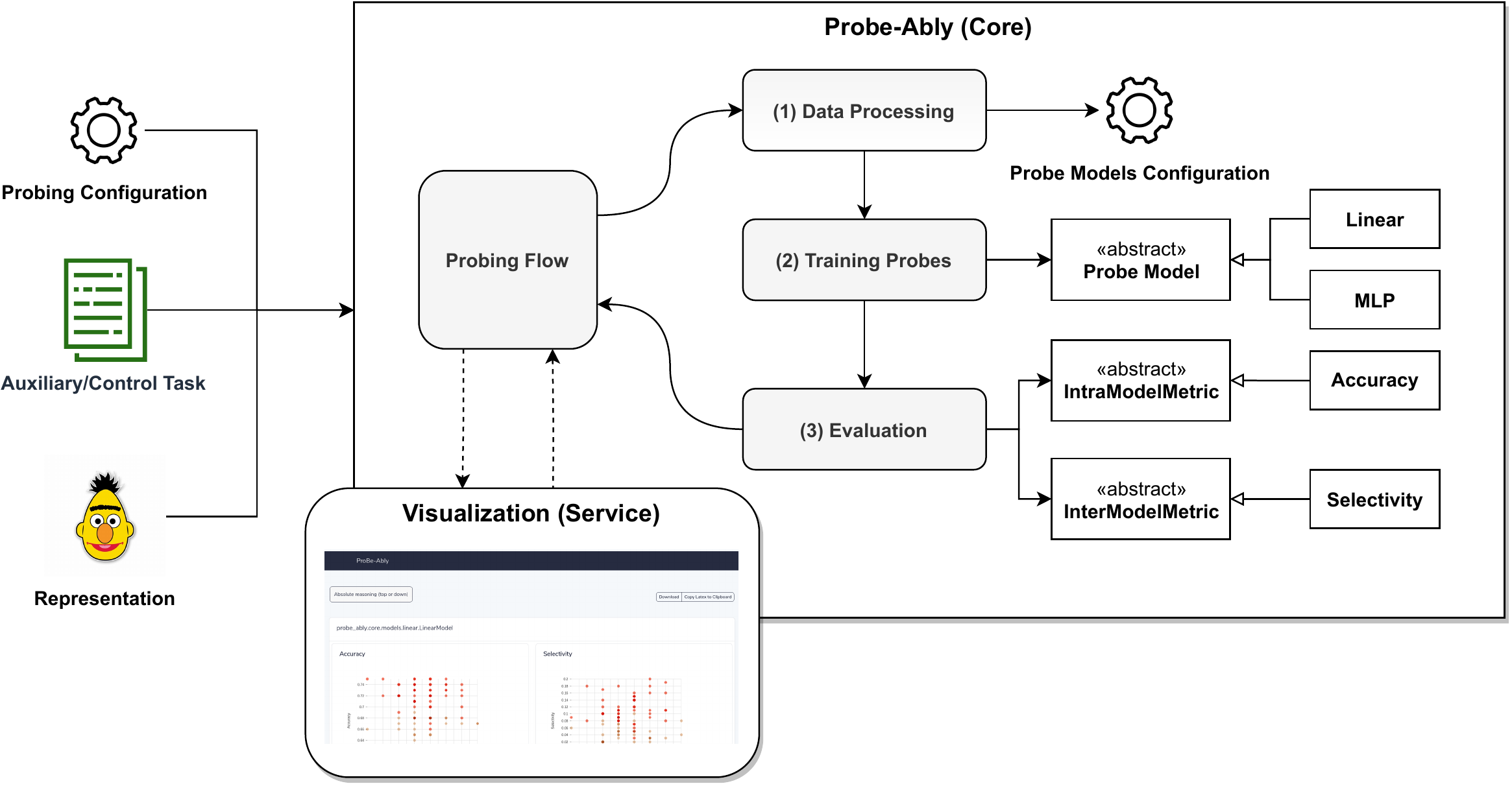}
        \caption{An overview of Probe-Ably. The core facility provided by Probe-Ably is the encapsulation of an end-to-end experimental probing pipeline. The framework offers a complete implementation and orchestration of the main tasks required for probing, together with a suite of standard probe models and evaluation metrics.}
        \label{fig:architecture}
\end{figure*}

Probe-Ably\footnote{Video demonstration: \\ \url{https://youtu.be/lE3O_BENBxk}} is a framework designed for PyTorch\footnote{\url{https://pytorch.org/}} to support researchers in the implementation of probes for neural representations in a flexible and extendable way.

The core facility provided by Probe-Ably is the encapsulation of the end-to-end experimental probing pipeline. Specifically, Probe-Ably provides a complete implementation of the core tasks necessary for probing neural representations, starting from the configuration and training of heterogeneous probe models, to the calculation and visualization of metrics for the evaluation.  

The probing pipeline and the core tasks operate on a set of abstract classes, making the whole framework agnostic to the specific representation, auxiliary task, probe model, and metrics used in the concrete experiments (see Fig \ref{fig:architecture}). This architectural design allows the user to:

\begin{enumerate}
    \item Configure and run probing experiments on different representations and auxiliary tasks in parallel;
    \item Automatically generate control tasks for the probing, allowing the computation of inter-model metrics such as \emph{selectivity};
    \item Extend the suite of probes with new models without the need to change the core probing pipeline;
    \item Customize, implement and adopt novel evaluation metrics for the experiments.
\end{enumerate}

\subsection{Probing Pipeline}

In this section we describe the core components implemented in Probe-Ably.

A probing pipeline is typically composed of the following sub-tasks:
\begin{enumerate}
    \item \textbf{Data Processing:} This task consists in data preparation and configuration of the probe models for the subsequent training task. For each representation to be probed and each auxiliary task, a requirement in this stage is the generation of a \emph{control task} \cite{hewitt-liang}, along with the selection of distinct hyperparameter configurations for the probe models. Generally, the control task can be either designed by researchers or automatically constructed by randomly assigning labels to the examples in the auxiliary task. On the other hand, the hyperparameter selection is crucial for the interpretation of the probing results, and has to guarantee a large coverage of the configuration space to allow for a significant comparison of the representations under investigation. Common methods for hyperparameter selection adopt a combination of grid search and random sampling techniques.  
    
    \item \textbf{Training Probes:} This task consists in training a set $\Phi$ of probe models. In particular, for each representation and each auxiliary task, researchers need to train probe models of different types (e.g., \emph{linear models}, \emph{multi-layer perceptrons}) and distinct hyperparameter configurations (e.g., hidden size, number of layers). Therefore, the number of probe models to be trained can rapidly increase with the number of representations, auxiliary tasks, and possible configurations.
    Let $n$ be the number of representations to be probed, $m$ the number of auxiliary tasks, $z$ the number of probe models, and $k$ the number of selected hyperparameter configurations for each probe. The total cardinality of $\Phi$ is generally equal to $|\Phi| = n \times m \times z \times k$. Thus, because of the potentially large space of models and configurations, the training task typically represents the most demanding and time-consuming stage in the overall probing pipeline. 
    
    \item \textbf{Evaluation:} The evaluation stage consists in calculating a set of metrics for assessing the performance and quality of the probes on the auxiliary tasks. The most common metrics adopted for probing evaluation are \emph{accuracy} and \emph{selectivity}. Generally, these quantities are plotted against the complexity of the probe models and are used to compare the trend in the performance of different neural representations on a given auxiliary task.  
\end{enumerate}

Probe-Ably provides a complete implementation and orchestration of the aforementioned tasks, which are integrated by a component named \emph{Probing Flow} (see Fig. \ref{fig:architecture}). 

The Probing Flow is ready to use for configuring and running standard probing experiments including hyperparameters selection via grid search. Moreover, the flow can be flexibly adapted to new models and metrics if necessary by extending the appropriate abstract classes and configuration files (additional details are described in section \ref{sec:customization}). We provide a pre-implemented suite of probe models and metrics whose details are described in sections \ref{sec:models} and \ref{sec:metrics}. 

In order to configure and run a new probing experiment, the user has to provide the following input:
\begin{itemize}
        \itemsep0em 
    \item \textbf{Probing Configuration:} a JSON file describing the components and parameters for the probing experiments. This file allows specifying the concrete probe models to train on each auxiliary task, along with pre-defined training parameters such as batch-size, number of epochs and number of different hyper-parameter configurations to test. Additionally, the probing configuration file can be used to indicate the metrics to use for the final evaluation.     
    \item \textbf{Auxiliary Task:} a TSV file containing the data and labels composing the auxiliary task. Probe-Ably allows the user to configure experiments that run on more than one auxiliary task in parallel. 
    \item \textbf{Control Task (Optional):} a TSV file containing the  labels composing a control task. The control tasks are automatically generated for each auxiliary task during the data processing stage. If not provided, we assign random labels to the example in the auxiliary tasksfor. 
    \item \textbf{Representation:} a TSV file containing the pre-trained embeddings for each example in the auxiliary task (e.g. BERT \cite{devlin2019bert}, RoBERTa \cite{liu2019roberta}). Similarly to the auxiliary tasks, Probe-Ably can run experiments on more than one representation in parallel. 
\end{itemize}

\begin{figure*}
        \centering
        \includegraphics[width=\textwidth]{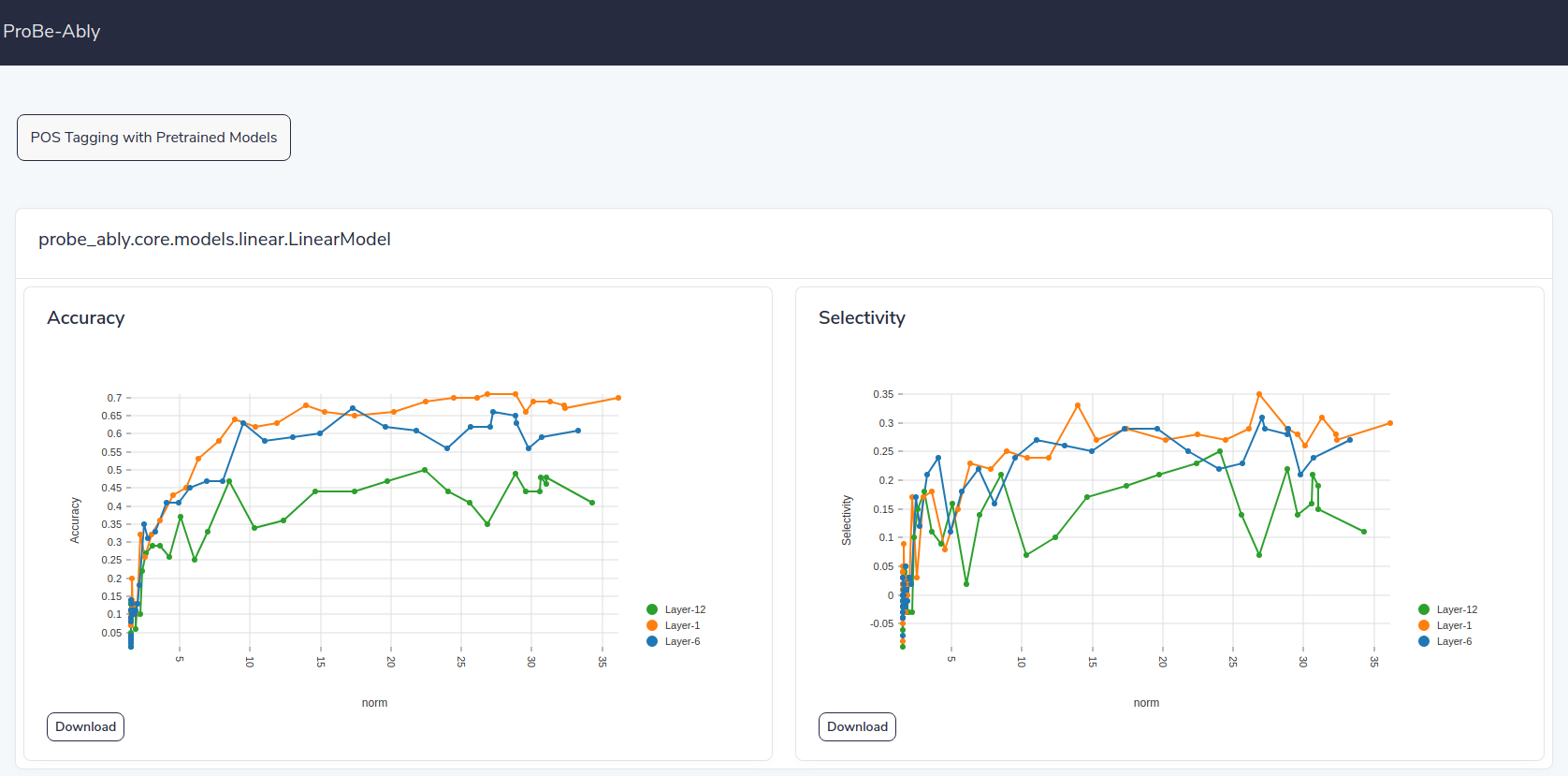}
        \caption{ Probe-Ably is integrated with a front-end visualization service, which supports researchers in consulting and plotting the results of their experiments.}
        \label{fig:visualization}
\end{figure*}

\subsection{Available Models}
\label{sec:models}
A common theme in probing studies is the use of structurally 
\emph{simple} classifiers: two common choices are \emph{linear} models and
\emph{multi layer perceptrons}\footnote{The hyperparameters of all implemented models are configurable, but we use the same default hyperparameter
ranges as \cite{pimentel2020information}.}.

Following works such as \cite{hewitt-manning} and 
\cite{pimentel2020pareto}, each instantiated model 
comes with some approximate appropriate \emph{complexity}.
This is varied in a controlled way in order to include results for a range 
of model complexities: this mitigates the possible confounding effect of 
\emph{overly expressive probes} which might be ``memorizing" the task
\cite{hewitt-liang, pimentel2020pareto}.

For linear models $ \hat{y} = W \mathbf{x} + \mathbf{b} $, 
 we mimic \cite{pimentel2020pareto} in using the nuclear norm 
$$ ||\mathbf{W}||_{*} = \sum_{i=1}^{min(|\mathcal{T}|, d)}\sigma_i(\mathbf{W}).$$ 
of the matrix $W$ as the approximate measure of complexity. 
The rationale here is that the nuclear norm approximates the rank of the transformation matrix. 
The rank may be used instead in situations where there is a large number of class labels, but as it is
limited by this number the nuclear norm presents a wider range of values.
The nuclear norm is included in the loss (weighted by a parameter $\lambda$) 
$$ 
-\sum_{i=1}^{n}\log p(t^{(i)} \mid \mathbf{h}^{(i)}) + \lambda \cdot ||\mathbf{W}||_{*}  $$
and is thus regulated in the training loop.

Multi-layer perceptrons are the only non-linear models currently included. 
Their flexibility and simplicity has made them popular choices in probing studies. 
We use the \emph{number of parameters} as an estimation of model complexity. 
Since sufficiently large MLP models could be prone to ``fitting" noise in the data,
it is especially important
to monitor the \emph{selectivity} when using this class of probes.

\subsection{Available Metrics}
\label{sec:metrics}
Certain probing metrics are not tied to the output of a specific probe, but to
two or more probes or training runs. 
As such, we have chosen to distinguish between \emph{intra-model} 
and \emph{inter-model} metrics.
\paragraph{Intra-Model Metrics.}
Individual model results and losses fall into this category. This includes the usual suspects
such as \emph{cross-entropy loss} and \emph{accuracy}.  
Intra-model metrics can be used for training, model-selection and reporting purposes.
\paragraph{Inter-Model Metrics.}
An important component of assessing the reliability of a probe's result is the 
\emph{selectivity} metric \cite{hewitt-liang}: 
for a fixed probe architecture and hyperparameter configuration, 
the auxiliary task accuracy is compared to the accuracy on a \emph{control task}, 
hence incorporating the results of two trained models. 
This is our primary example of an inter-model metric, but this format could be useful for other probing metrics such as minimum description length (online code version) 
\cite{voita2020information} or pareto hypervolume \cite{pimentel2020pareto}, 
which incorporate the results of multiple models or training runs.
These are only used for \emph{reporting} purposes, as they are external to each model's training
loop.
         
\subsection{Front-end Visualization}


Probe-Ably is integrated with a front-end visualization service. 
The front-end is used to plot the results of each probing experiment in a user-friendly way. 
The service is designed to be accessible via standard web browsers, and support researchers in analysing and comparing the probing performance of each representation on different auxiliary tasks. 

An example of plots included in the front-end visualization is shown in Figure \ref{fig:visualization}. Each plot can be downloaded in a pdf format to be stored locally or integrated in a LaTeX project. 

\section{Customized Probing Experiments}
\label{sec:customization}

Probe-Ably can be flexibly adapted and extended to run experiments on different representations, novel probe models and evaluation metrics. The following sections provide an overview of how researchers and users can customize their experiments via configuration files or implementation of new concrete classes. 

For a complete guide on how to extend and customize Probe-Ably, please consult the documentation\footnote{Documentation:\\ \url{https://ai-systems.github.io/Probe-Ably/}}\footnote{Repository:
\url{https://github.com/ai-systems/Probe-Ably/}}.

\subsection{Configuration}
Although default configurations are ready to use to run a basic set of experiments, the details of the latter can be customized according to specific needs, using the apposite probing configuration file. This pertains to aspects such as probe model choice, number of experiments, auxiliary tasks labels, input representations and custom control labels. 

Therefore, the settings can be modified by providing or editing the values of the attributes in the configuration file which specifies details about auxiliary tasks, probing model/s and training regime, including paths to any custom metrics or models. 

The structure of the probing configuration file is as follows:

\begin{itemize}
    \item tasks (list)
    \begin{itemize}
        \item task\_name (attr)
        \item representations (list)
        \begin{itemize}
            \item representation\_name (attr)
            \item file\_location (attr)
            \item control\_location (attr)
        \end{itemize}
    \end{itemize}
    \item probing\_setup (dict)
    \begin{itemize}
        \item train\_size (attr)
        \item dev\_size (attr)
        \item test\_size (attr)
        \item intra\_metric (attr)
        \item inter\_metric (attr)
        \item probing\_models (list)
        \begin{itemize}
            \item probing\_model\_name (attr)
            \item batch\_size (attr)
            \item epochs (attr)
            \item number\_of\_models (attr)
        \end{itemize}
    \end{itemize}
\end{itemize}

\subsection{Adding a Probe Model}
\label{sec:add_models}
Custom probe models can be introduced by extending the abstract \texttt{ProbeModel} class (Fig. \ref{fig:architecture}). This class inherits the methods and attributes of a \texttt{nn.Module} in PyTorch.
To extend Probe-Ably with a new probe model, the user needs to implement two methods, namely \texttt{forward} and \texttt{get\_complexity}.

The \texttt{forward} method is inherited from PyTorch and is adopted to compute the predictions of the probe models along with their loss function. On the other hand, the \texttt{get\_complexity} method has to return a complexity measure for the model (e.g., nuclear norm, number of parameters). This method is internally used by the Probing Flow for setting up and executing the probing pipeline, and creating the right visualization for the results. 

In order to make a customized probe model available for new experiments, the user needs to specify a model configuration file (JSON format) containing the path to the concrete class, together with the parameters required for its instantiation. The model configuration file is organized as follows:

\begin{itemize}
    \item model\_class (attr)
    \item params (list)
    \begin{itemize}
        \item name (attr)
        \item type (attr)
        \item options (attr)
    \end{itemize}
\end{itemize}

\subsection{Adding an Evaluation Metric}
\label{sec:add_metrics}

Similarly to probe models, it is possible to extend Probe-Ably with new evaluation metrics. In order to add a new metric, the user can extend one of the available abstract classes (i.e., \texttt{IntraModelMetric} or \texttt{InterModelMetric}).

In this case, it is not necessary to specify a configuration file for the metrics, and the user only needs to implement the apposite function, \texttt{calculate\_metrics}, that performs the appropriate computation. Subsequently, the user can adopt the new metric in a probing experiment by editing the apposite attribute in the probing configuration file.

\begin{figure*}[ht]
    \centering
    \subfloat[Linear Model Accuracy.\label{fig:acc_linear}]{\includegraphics[width=0.8\textwidth]{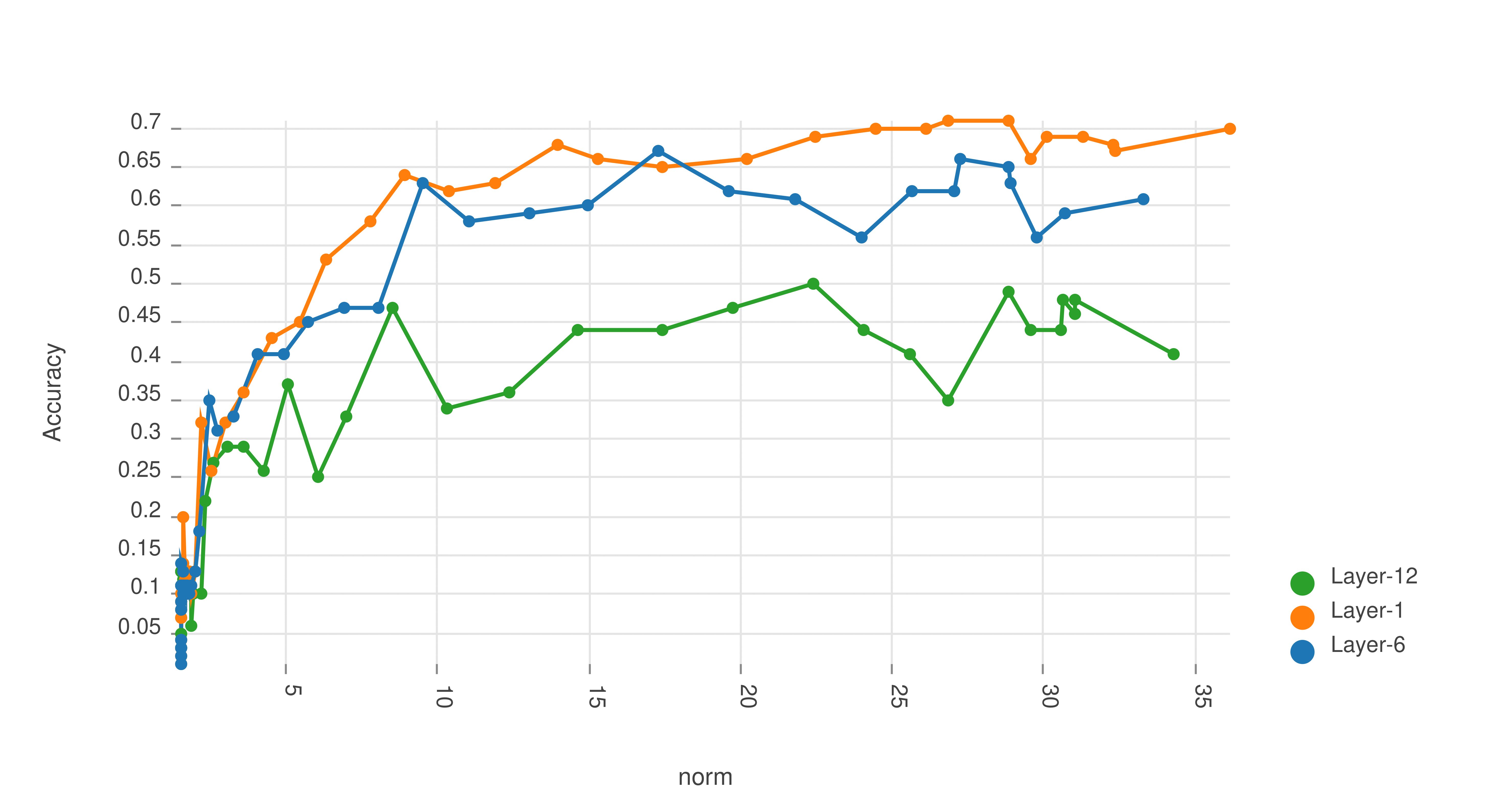}}
    \\
    \subfloat[Linear Model Selectivity.\label{fig:sel_linear}]{\includegraphics[width=0.8\textwidth]{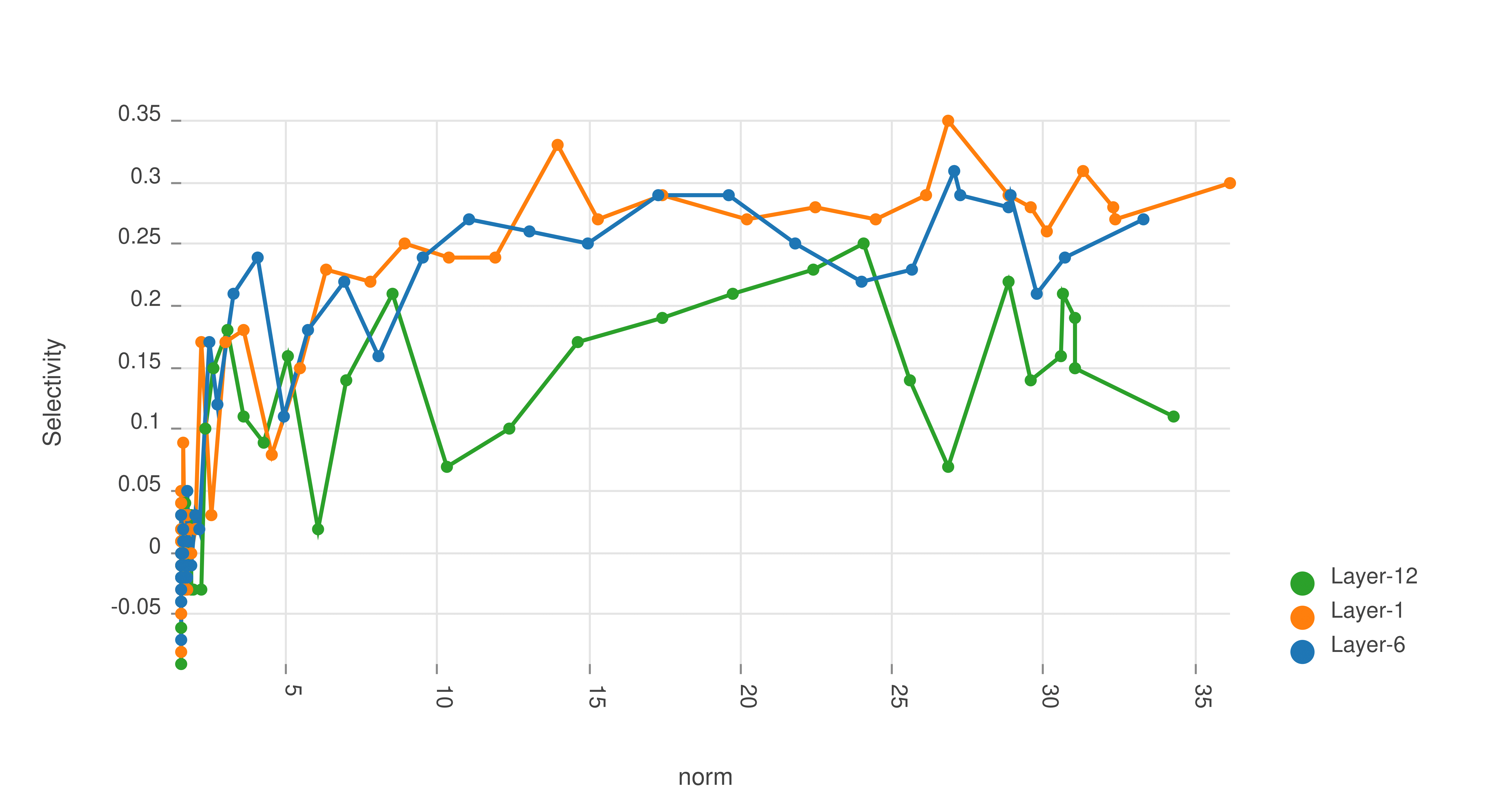}}
    \caption{Probing results for different layers of
BERT on the \emph{Part-Of-Speech} task using the control task presented in \cite{hewitt-liang}, implemented and executed through Probe-Ably (see Section \ref{sec:case_study}). The results are consistent with observations 
in \cite{tenney2019bert}, which note that syntactic features (such as part
of speech tags) are more prevalent in earlier layers of BERT. }
\end{figure*}

\section{Interpreting Results}
We provide the following list of guidelines for interpreting results:
\begin{itemize}
    \itemsep0em 
    \item Regions of low selectivity indicates a less trustworthy auxiliary task accuracy result. As accuracy increases with model complexity, keep an eye on the selectivity value: if it starts to drop again, this 
    indicates that the probe is expressive enough to fit the randomized control task 
    (and thus high expressivity and overfitting may be responsible for a high auxiliary task accuracy).
    \item We recommend a focus on \emph{comparison of trends} between models/representations
        rather than probe performance on any fixed set of representations.
    \item These comparisons are more convincing if they are consistent across a range of probe complexities.  
    \item Note that any given probe architecture imposes a structural assumption.
    For example, \emph{linear} probes may only attain a high accuracy if the representation-target
    relationship is linear. We recommend that these assumptions/probe model choices be guided by
    prior visualizations and hypothesized relationships.
    \item As far as possible, stick to comparing representations of the same sizes. Lower-dimensional representations may reach their maximum accuracy at lower probe complexity values; 
    as such they may give the ``appearance" of superior probe accuracy scores to 
    larger representations.
    For this reason, it is also important that you investigate a sufficiently large range of
    model complexities.
 \end{itemize}
 
\section{Case Study}
\label{sec:case_study}
To demonstrate the Probe-Ably system, we include an implementation of a Part-Of-Speech tagging
auxiliary task based on the Penn Treebank corpus \cite{marcus}. It has been used multiple times in works on probing methodology \cite{hewitt-liang, voita2020information, pimentel2020information}.
We use the custom control task from \cite{hewitt-liang}.
Using linear models as probes, we compare the probing results for different layers of
BERT (\texttt{bert-base-uncased}) pre-trained on the masked language modelling task \cite{devlin2019bert},
across 50 probing runs.
The results are consistent with observations 
in \cite{tenney2019bert}, which note that syntactic features (such as part
of speech tags)
are more prevalent in earlier layers of BERT. 
This case study is available as a ready-to-run example. 

\section{Related Work}
Previous interpretability tools for neural models have focused on gradient-based methods 
\cite{Wallace2019AllenNLP}, the visualization of attention weights \cite{vig} and other tools
focusing on NLP model explainability and interpretability \cite{what-if, tenney2020language}.

The ongoing discussion on probing, auxiliary tasks and the surrounding best practices can be 
traced back to
the early definitions in \cite{alain-bengio}, where it was first described as 
\emph{diagnostic classification}. Early probing studies in NLP include \cite{zhang-bowman} and 
\cite{tenney}, the former being an early example of the importance of comparing with
randomized representations or labels.
Further discussion has introduced control tasks and the selectivity metric \cite{hewitt-liang},
formalized notions of \emph{ease of extraction} \cite{voita2020information} and
described other strategies for taking model complexity into account 
\cite{pimentel2020pareto}. 

\section{Conclusion}
While probing can be used to explore hypotheses about linguistic (or general)
features present in model representations, there are various pitfalls that can 
lead to premature or incorrect claims. 
Much progress has been made in establishing better practices for probing studies,
but these involve running large systematic sets of experiments employing recently-developed metrics and correctly interpreting results. \textbf{Probe-Ably} is designed to simplify and encourage the use of 
emerging methodological developments in probing studies, serving as a task-agnostic and model-agnostic platform 
for auxiliary diagnostic classification for high-dimensional vector representations.

\section*{Acknowledgements}

The authors would like to thank the anonymous reviewers for the constructive feedback.  Additionally, we would like to thank the Computational Shared Facility of the University of Manchester for providing the infrastructure to run our experiments. Thanks to Alber Santos for the helpful discussions.

\bibliographystyle{acl_natbib}
\bibliography{acl2021}


\end{document}